\documentclass[letterpaper, 10 pt, conference]{ieeeconf}  % Comment this line out if you need a4paper

\IEEEoverridecommandlockouts                              % This command is only needed if 
\usepackage{graphics} % for pdf, bitmapped graphics files
\usepackage{epsfig} % for postscript graphics files
\usepackage{mathptmx} % assumes new font selection scheme installed
\usepackage{times} % assumes new font selection scheme installed
\usepackage{amsmath} % assumes amsmath package installed
\usepackage{amssymb}  % assumes amsmath package installed
\usepackage{float}
\usepackage{graphicx}
\usepackage{subcaption}
\usepackage{caption}
\usepackage{xcolor}
\usepackage{url}
\usepackage{cuted}
\usepackage{capt-of} 
\usepackage{nicefrac}
\usepackage{booktabs}
\usepackage[scale=1.1]{inconsolata} % nice monospaced font, slightly smaller

\usepackage{hyperref}
\definecolor{linkcolor}{HTML}{328DD3}
\hypersetup{
    colorlinks=true,
    linkcolor=black,
    citecolor=black,
    urlcolor=linkcolor
}
\makeatletter
\let\NAT@parse\undefined
\makeatother
\usepackage[numbers,sort,compress]{natbib}

\title{\LARGE \bf
Masquerade: Learning from In-the-wild Human Videos using Data-Editing}

\author{
    Marion Lepert$^{\ast, 1}$, Jiaying Fang$^{\ast, 1}$, Jeannette Bohg$^{1}$\\[0.1cm]
    $^{1}$Stanford University\\[2pt]
    \small\href{https://masquerade-robot.github.io}{%
        \ttfamily\textmd{https://masquerade-robot.github.io}%
    }\\[2pt]
    \vspace{-1cm}
    \thanks{$^\ast$: Equal contribution. \newline
    \texttt{marion.lepert@gmail.com}, \texttt{\{jyfang, bohg\}@stanford.edu}}
}

\begin{document}

\maketitle
\thispagestyle{empty}
\pagestyle{empty}

\begin{strip}
  \centering
  \includegraphics[width=\textwidth]{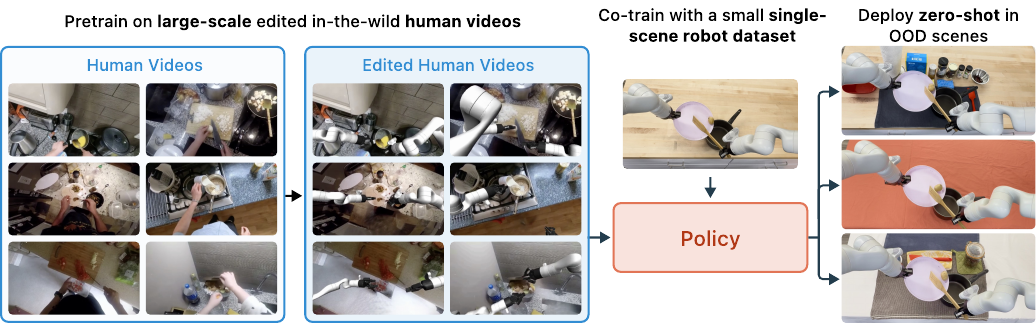}
  \captionof{figure}{Overview of Masquerade. \textbf{Left}: Large‑scale in‑the‑wild egocentric human videos are edited to obtain “robotized” demonstrations that bridge the visual embodiment gap. A vision representation is pre‑trained to predict future 2D robot poses on 675K frames of these edited clips. \textbf{Center}: the vision representation is co-trained with a diffusion policy head on 50 real robot demonstrations collected in a single scene. \textbf{Right}: The resulting policy is deployed zero‑shot in previously unseen environments, achieving significantly more robust manipulation performance than baselines despite domain shifts.}
  \label{fig:intro}
\end{strip}

%%%%%%%%%%%%%%%%%%%%%%%%%%%%%%%%%%%%%%%%%%%%%%%%%%%%%%%%%%%%%%%%%%%%%%%%%%%%%%%%
\begin{abstract}
Robot manipulation research still suffers from significant data scarcity: even the largest robot datasets are orders of magnitude smaller and less diverse than those that fueled recent breakthroughs in language and vision. We introduce Masquerade, a method that edits in-the-wild egocentric human videos to bridge the visual embodiment gap between humans and robots and then learns a robot policy with these edited videos. Our pipeline turns each human video into ``robotized" demonstrations by (i) estimating 3‑D hand poses, (ii) inpainting the human arms, and (iii) overlaying a rendered bimanual robot that tracks the recovered end‑effector trajectories. Pre‑training a visual encoder to predict future 2‑D robot keypoints on 675K frames of these edited clips, and continuing that auxiliary loss while fine‑tuning a diffusion‑policy head on only 50 robot demonstrations per task, yields policies that generalize significantly better than prior work. On three long‑horizon, bimanual kitchen tasks evaluated in three unseen scenes each, Masquerade outperforms baselines by 5-6×. Ablations show that both the robot overlay and co‑training are indispensable, and performance scales logarithmically with the amount of edited human video. These results demonstrate that explicitly closing the visual embodiment gap unlocks a vast, readily available source of data from human videos that can be used to improve robot policies.
\end{abstract}

%%%%%%%%%%%%%%%%%%%%%%%%%%%%%%%%%%%%%%%%%%%%%%%%%%%%%%%%%%%%%%%%%%%%%%%%%%%%%%%%
\section{Introduction}
Recent successes in natural language processing (NLP) and computer vision (CV) stem from training on massive, diverse datasets. In robotics, however, data scarcity remains a major bottleneck: collecting real-world robot data is slow and expensive, so even the largest robotics datasets are orders of magnitude smaller than those in NLP/CV. As a result, generalist robot policies still lag far behind their language and vision model counterparts.

Human videos provide a rich supplement to limited robot datasets, spanning countless real-world manipulation scenarios at massive scale. However, leveraging human videos for robot policy learning is challenging because human videos lack precise action labels and feature an inherent embodiment gap: humans look and move differently from robots. Prior works have addressed these problems by training on human videos using proxy tasks, such as pre-training vision encoders \cite{nair2023r3m, karamcheti2023voltron, radosavovic2023real, majumdar2023we, srirama2024hrp}, inferring reward functions \cite{chen2021learning, mavip, bhateja2024robotic}, or learning world models \cite{wu2023unleashing, du2023learning, bharadhwaj2024gen2act, cheang2024gr}. These works typically do not explicitly address the visual embodiment gap between humans and robots, and instead assume that the model will implicitly learn correspondences between the human and robot embodiments by training on both types of data.

In this work, we ask whether explicitly closing that gap—even imperfectly—can unlock more signal from human videos. We extend Phantom’s \cite{lepert2025phantom} data-editing pipeline—which was demonstrated only on carefully collected single-hand human video demonstrations with a fixed camera—to in-the-wild videos. Specifically, we estimate hand poses, inpaint away the human body, render a simulated robot in the same pose, and overlay it back into each frame. This yields a large, “robotized” video dataset.

We then follow the now-standard recipe of broad pretraining followed by focused finetuning: we pretrain our vision encoder to predict future 2D robot poses on the edited videos, and subsequently co-train this vision encoder with a policy head on a small set of real robot demos in a single scene. We find that retaining the pretraining objective is crucial during finetuning to obtain out-of-distribution robustness in novel scenes.

Across three challenging bimanual tasks and three novel environments each, our method produces policies that generalize far beyond baselines. \textbf{Our key contribution is showing that explicitly addressing the visual embodiment gap between humans and robots—even via simple 2D overlays— substantially enhances what robot policies can learn from in-the-wild human videos.}

\section{Related Works}
Robotics research has increasingly turned to human video data as a way to overcome the scarcity of robot demonstrations. Such videos come in two broad forms:

\textbf{In-the-wild human videos} — uncurated internet videos of people performing everyday, unscripted activities in diverse environments, often with occlusions, and camera motion. These videos offer massive scale and diversity but lack robot-friendly data quality or precise action labels.

\textbf{Curated human video demonstrations} — videos intentionally recorded for robot learning, with task-focused motions, minimal occlusions, and often captured with specialty hardware such as depth cameras or AR/VR devices to provide accurate hand pose annotations.

\subsection{Learning from In-the-wild Human Videos}
A growing body of work seeks to leverage in-the-wild human videos to bootstrap robotic learning. One line of work pretrains visual encoders on human videos for downstream tasks. R3M \cite{nair2023r3m} uses time-contrastive and video–language objectives on Ego4D \cite{grauman2022ego4d}, while Voltron \cite{karamcheti2023voltron} aligns video with captions via reconstruction and generation losses. Masked auto-encoding approaches like MVP \cite{radosavovic2023real} and VC-1 \cite{majumdar2023we} adapt MAE transformers \cite{he2022masked} to human clips. HRP \cite{srirama2024hrp} extracts affordance signals—future contact points, hand poses, and objects—and pretrains a vision backbone on these self-supervised tasks.  

Beyond representation learning, a second line of work has leveraged in-the-wild human videos to provide rich auxiliary supervision for downstream robotic tasks. For example, several methods infer reward functions directly from video demonstrations \cite{Shao-RSS-20,chen2021learning, mavip, bhateja2024robotic}, while others learn predictive world models by training dynamics encoders on raw video data \cite{wu2023unleashing, du2023learning, bharadhwaj2024gen2act, cheang2024gr}. Another group of approaches extracts hand-pose trajectories from human clips to derive motion priors for robot policies \cite{shaw2023videodex, mandikal2022dexvip, singh2024hand, kannan2023deft, bharadhwaj2024towards, bharadhwaj2024track2act}, and yet another direction focuses on discovering object-centric affordances—mapping how objects should move from human videos \cite{bahl2023affordances, chen2025vidbot, shi2025zeromimic}. LAPA \cite{yelatent} learns discrete latent actions from human videos via a VQ-VAE \cite{van2017neural} objective and uses these latents to fine-tune a VLA on small-scale robot data. 

Despite these advances, none of these works explicitly address the large visual embodiment gap between human hands and robot grippers, making it challenging for vision-based policies—often brittle to out-of-distribution appearance shifts—to transfer learned representations from human videos to robots.  Our method directly closes this gap through simple 2D inpainting of human hands into robot grippers, and we find that even this imperfect visual alignment yields surprisingly large gains in cross-embodiment transfer. Concurrent, unpublished work H2R \cite{li2025h2r} also uses a Phantom-like pipeline \cite{lepert2025phantom} but relies solely on finetuning—a strategy we show to be markedly less effective—and reports only minor gains on simple tasks. In contrast, we pair closing the embodiment gap with a co-training pipeline that effectively leverages edited in-the-wild human videos, enabling robust performance on challenging, long-horizon bimanual tasks.

\subsection{Learning from Curated Human Video Demonstrations}
To overcome the lack of ground‐truth actions in raw in-the-wild human videos, many methods focus instead on learning from curated human video demonstrations. These videos contain clean, task-related human motions with minimal occlusions or camera motion. Some works leverage these more accurate action labels and propose treating humans as another robot embodiment and co-training policies on human and robot data \cite{kareer2024egomimic, qiu2025humanoid, yang2025egovla}. EgoMimic \cite{kareer2024egomimic} and PH$^{2}$D \cite{qiu2025humanoid} jointly train on egocentric human demonstrations (captured with a wearable camera) and teleoperated robot trajectories via a shared vision–policy backbone and cross-domain alignment losses. EgoVLA \cite{yang2025egovla} trains a pretrained vision-language model on both human and robot data.

Other approaches learn implicit motion priors from curated human video demonstrations \cite{zakka2022xirl, wang2023mimicplay, xu2023xskill, qin2022dexmv}. Several works leverage object-centric trajectories or point flows \cite{zhu2024vision, heppert2024ditto, xu2024flow, haldar2025point, ren2025motion, lum2025crossinghumanrobotembodimentgap} to bridge the visual embodiment gap between humans and robots. Other methods \cite{bahl2022human, duan2023ar2, lepert2025phantom} use inpainting. Whirl \cite{bahl2022human} collects human demonstrations on multiple tasks, and then inpaints out human hands in human demonstrations and robot arms in robot demonstrations to bridge the visual gap. Phantom \cite{lepert2025phantom} learns policies zero-shot from human videos by inpainting out human hands and overlaying simulated robot arms on observation images. 

While minimizing the visual embodiment gap and co-training on human and robot data have proven effective on hand collected datasets, their application to large-scale, in‐the‐wild internet videos, which offer far greater scale and diversity, remains unexplored. In this work, we show that the combination of both techniques can be successfully extended to in-the-wild human videos to obtain more robust policies. Compared with only using curated human video demonstrations, this enables using significantly larger human video datasets for robot learning.

\section{Method}
\subsection{Problem Setup}

\begin{figure*}
    \centering
    \includegraphics[width=\textwidth]{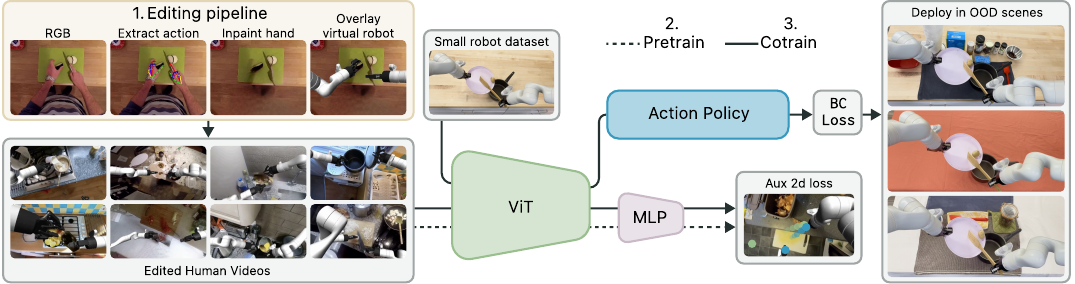}
    \caption{Overview of Masquerade. (1) In-the-wild egocentric human videos are converted into “robotized” clips by extracting 2D hand poses, inpainting out the human arms, and overlaying a rendered bimanual robot in the same pose. (2) A ViT-Base vision encoder is pretrained on these edited videos using a 2D keypoint regression loss. (3) During cotraining, the encoder and a diffusion-based policy head are jointly optimized on a mix of edited human videos (auxiliary 2D loss) and real robot demonstrations (imitation loss).}
    \label{fig:method}
\end{figure*}

We assume access to a large‐scale in-the-wild human video dataset $\mathcal{D}_{\mathrm{human}}=\{\tau^{(h)}_i\}_{i=1}^N$ where each $\tau^{(h)}_i$ is a human video clip. We use the Epic Kitchens dataset \cite{Damen2018EPICKITCHENS}, which contains a wide range of naturally occurring bimanual kitchen tasks recorded in diverse real-world scenes with egocentric cameras. These videos capture people performing their normal, unscripted everyday activities, with no effort to make the content more suitable for robot learning. For each clip, we also have an associated natural language annotation describing the activity depicted.

We additionally have a small set of bimanual robot demonstrations of a given task $\mathcal{D}_{\mathrm{robot}}=\{\tau^{(r)}_j\}_{j=1}^M$ captured from the robot’s egocentric camera, with known intrinsics and extrinsics.

Our method proceeds in three stages: first, we edit the human video dataset to reduce the human-to-robot embodiment gap; next, we pretrain a vision encoder on the edited human videos to learn rich, in-the-wild features; and finally, we co-train an imitation learning policy on robot data alongside the edited human data to transfer these learned priors.

\subsection{Data processing of in-the-wild egocentric videos}

Human videos pose two main challenges for robot policy learning: a large visual embodiment gap and missing action labels. We address these by using a modified version of the Phantom \cite{lepert2025phantom} pipeline to convert each human clip into a synthetic robot demonstration and then extracting 2D hand keypoints in each frame to use as action labels.

\subsubsection{Visual editing of in-the-wild videos}

Let each human demonstration $\tau^{(h)}_i$ be a sequence of egocentric frames \(\{I_{t}^{(h)}\}_{t=1}^T\). We localize the left and right hands in each frame using the Epic Kitchens annotations and estimate 21 anatomical keypoints per hand with HaMeR \cite{pavlakos2024reconstructing}. These keypoints \(\hat{\mathbf X}_t\in\mathbb R^{21\times3}\) are mapped to a 3D robot end‐effector pose \(\mathbf P_t=(\mathbf p_t,\mathbf R_t,g_t)\) following \cite{lepert2025phantom}, where $\mathbf{p}_t \in \mathbb{R}^3$ is the Cartesian position, $\mathbf{R}_t$ is the orientation, and $g_t \in [0, 1]$ is the normalized gripper opening width. The poses $\mathbf P_t$ are temporally smoothed to reduce noise.

Next, we segment out human arms using Detectron2 \cite{wu2019detectron2} and SAM2 \cite{ravisam} and remove them via E2FGVI inpainting \cite{liCvpr22vInpainting}. Using the known camera intrinsics and extrinsics, we render a virtual bimanual robot model whose end effectors follow \(\mathbf P_t\), and composite this render into the original view. The result is a video that appears to show the robot performing the task (see Fig.~\ref{fig:method}). All edited clips form our modified dataset \(\mathcal{D}_{\mathrm{human}}'\). 

\subsubsection{Extracting training labels from in-the-wild videos} 

Although HaMeR reliably recovers hand shape and 2D keypoint locations, its monocular input precludes accurate absolute 3D pose estimation. Unlike \cite{lepert2025phantom}, which refines HaMeR with depth, our large-scale human videos lack depth data. 
Therefore, we use the 2D keypoint locations as supervisory labels for an auxiliary loss in our vision model, without incorporating them directly into policy learning.
To obtain the labels, we project the temporally smoothed 3D end‐effector positions \(\mathbf{p}_t\) onto the image plane—using known intrinsics and extrinsics—to obtain 2D action waypoints $\mathbf{p}_{t,2\mathrm{D}} \in \mathbb{R}^2$. Rather than supervising on only the next waypoint, we provide the encoder with a sequence of the next \(H\) waypoints as the prediction target:
\begin{align}
\mathbf{p}_{t:t+H,2\mathrm{D}} = (\mathbf{p}_{t,2\mathrm{D}}, \mathbf{p}_{t+1,2\mathrm{D}},...,\mathbf{p}_{t+H,2\mathrm{D}})   
\end{align}

To correct for egocentric camera motion, we compute a homography from frame $t$ to each future frame and warp all subsequent keypoints back into frame $t$’s view before forming this sequence. 

\subsubsection{Data Filtering}
Even after compensating for camera motion via homographies, excessive camera movement remains undesirable for our fixed‐base robot with a statically mounted camera. We therefore filter out frames where the estimated camera motion exceeds a threshold, as well as frames where the extracted actions are invalid due to keypoint errors or kinematic limits. This ensures that only stable, reliably labeled clips are used for policy learning.

While this filtering removes the most problematic cases, many overlays remain imperfect. Our retargeting pipeline cannot handle all dexterous grasps seen in in-the-wild videos, and the absence of depth data prevents correct handling of occlusions, sometimes causing robot pixels to erroneously appear over scene objects. Nevertheless, we show that these imperfect overlays dramatically improve performance compared to using no overlays at all —highlighting how even rough visual alignment can strongly benefit cross-embodiment transfer.

\subsection{Policy learning}
Our architecture consists of a language-conditioned vision encoder \(f(x,z)\) and a diffusion‐based action head \(g(\cdot)\). 

\subsubsection{Vision encoder pretraining}: We first pretrain \(f\) on our processed human dataset \(\mathcal{D}_{\mathrm{human}}'\) using 2D action supervision, conditioning the encoder on the per-clip language annotations via FiLM \cite{perez2018film}. Each language embedding is applied to all frames within its corresponding clip, allowing the encoder to modulate visual features based on the high-level semantic description of the activity. Concretely, we minimize
\[
\mathcal{L}_{\mathrm{2D}}
=\bigl\lVert h\Big(f(x, z_x\big)\Big) - \mathbf{p}_{t:t+H,2\mathrm{D}}\bigr\rVert^2,\quad
x \sim \mathcal{D}_{\mathrm{human}}'
\]

% \[
% \mathcal{L}_{\mathrm{2D}}
% =\bigl\lVert h\Big(\mathrm{FiLM}\big(f(x), z_x\big)\Big) - \mathbf{p}_{t:t+H,2\mathrm{D}}\bigr\rVert^2,\quad
% x \sim \mathcal{D}_{\mathrm{human}}'
% \]

where \(h\) is a small MLP that maps encoder features to 2D keypoint targets \(\mathbf{p}_{t:t+H,2\mathrm{D}}\) and $z_x \in \mathbb{R}^d$ is a fixed per clip language embedding associated with frame $x$. 

\subsubsection{Policy learning using cotraining}
Next, we train an imitation learning policy using a small set of task-specific robot demonstrations $\mathcal{D}_{\mathrm{robot}}$. We continue to optimize the pre-training loss with respect to the edited human videos during this training. To minimize the visual gap between $\mathcal{D}_{\mathrm{human}}'$ and $\mathcal{D}_{\mathrm{robot}}$, we inpaint a rendered robot over the robot so that the model is always seeing an inpainted robot. 

During co-training, we introduce a second loss: 

\[
\mathcal{L}_{\mathrm{policy}}
=\bigl\lVert g\bigl(f(y)\bigr) - \mathbf{P}^{(r)}\bigr\rVert^2,\quad
y \sim \mathcal{D}_{\mathrm{robot}}
\]

where $\mathbf{P}^{(r)}$ is the robot Cartesian end-effector action from the real robot data. We train both losses simultaneously

\[
\mathcal{L}
=  \mathcal{L}_{\mathrm{2D}} + \lambda \mathcal{L}_{\text{policy}}  \]

where \(\lambda\) is a hyperparameter chosen empirically. 

\begin{figure*}
    \centering
    \includegraphics[width=1\textwidth]{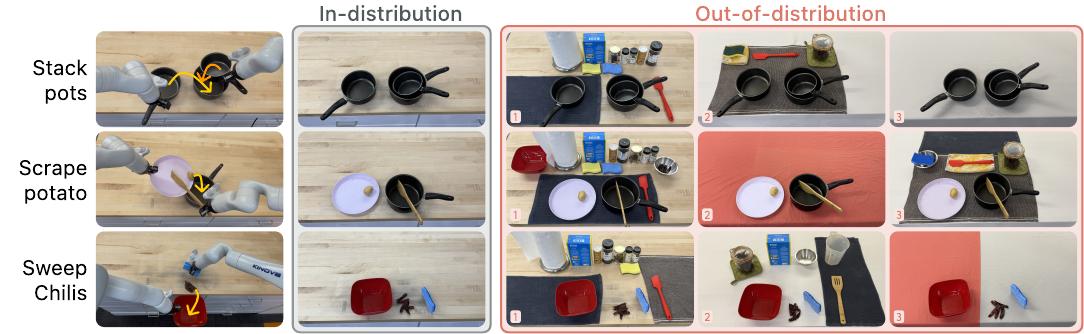}
    \caption{Scenes used for each task in in-distribution (center) versus out-of-distribution (right) settings; the first row represents the Stack Pots scenes, the middle the Scrape Potato scenes, and the bottom row the Sweep Chilis scenes.}
    \label{fig:scenes}
\end{figure*}

\section{Results}
We evaluate our method on three challenging bimanual tasks using a dual‐Kinova‐arm setup (see Fig. \ref{fig:hardware}). Our policy is trained on 10K clips (675K frames) from Epic Kitchens \cite{Damen2018EPICKITCHENS} and 50 task‐specific robot demonstrations collected in a single scene. For each clip in \(\mathcal{D}_{\mathrm{human}}'\), we generate a fixed embedding of the natural language video description using DistilBERT \cite{sanh2019distilbert}. The vision encoder $f(x)$ is a ViT-Base network initialized with ImageNet weights \cite{dosovitskiy2020image,imagenet15russakovsky,he2022masked}, and the action head follows the Diffusion Policy architecture \cite{chi2023diffusionpolicy}.

\subsection{Task descriptions}
We evaluate on three long-horizon bimanual tasks in out-of-distribution scenes shown in Fig.~\ref{fig:scenes} (see  Fig.\ref{fig:task_variations} for additional details). Because our tasks are long horizon, we capture partial progress in each rollout, by assigning each subtask one third of the total score:

\textbf{Stack Pots}
\begin{enumerate}
  \item Lift the small pot out of the large pot
  \item Insert the medium pot into the large pot
  \item Place the small pot inside the medium pot
\end{enumerate}

\textbf{Scrape Potato}
\begin{enumerate}
  \item Lift the plate carrying the potato
  \item Lift the spatula
  \item Scrape the potato into the pot using the spatula
\end{enumerate}

\textbf{Sweep Chilis}
\begin{enumerate}
  \item Grab the bowl and move it to the edge of the table
  \item Pick up the sponge
  \item Sweep the chilis into the bowl 
\end{enumerate}

\subsection{Baselines}
Our experiments are designed to answer the following question: \textbf{does editing in-the-wild human videos before using them for policy learning improve robot performance?} We compare against several baselines to directly probe this question. Because Masquerade leverages human videos to improve the policy’s vision representation, we focus our comparisons on (i) a state-of-the-art vision representation learned from human videos, and (ii) the most widely used general-purpose vision representations in robotics. 

\textbf{HRP} \cite{srirama2024hrp}: Finetunes a vision encoder on 150K egocentric human video clips (number of frames not reported) by regressing three affordance labels—future hand pose, active-object bounding box, and contact-point locations—automatically mined from raw videos. The resulting encoder is then used to train an imitation-learning policy. Notably, this work uses raw human videos and does not continue to co-train the vision encoder on human videos during policy learning. We use open sourced model weights. 

\textbf{ImageNet}: A ViT initialized on ImageNet-1K \cite{imagenet15russakovsky} remains one of the most reliable backbones for robot control. In an unbiased, rigorous study done by \cite{dasari2023datasets}, ImageNet pretraining outperformed robotics-specific human-video pretrained representations, including R3M \cite{nair2023r3m}, VC-1 \cite{majumdar2023we}, and MVP \cite{radosavovic2023real}.

\textbf{DINOv2} \cite{oquab2023dinov2}: DINOv2 is a high-capacity, self-supervised ViT trained on 142M curated images that yields strong general-purpose features and is also widely adopted in robotics \cite{kim2024openvla, bu2025univla, qiu2025humanoid, yang2025egovla, lin2024data, kerr2024robot, xia2024cage, yang2024harmonic, fang2025airexo, huang2024rekep, jia2024lift3d}. We include DINOv2 as a competitive, modern baseline.

All models, including ours, use the ViT-base architecture \cite{dosovitskiy2020image}.

\subsection{Performance in OOD scenes}
\begin{figure}
    \centering
     \includegraphics[width=1.05\linewidth]{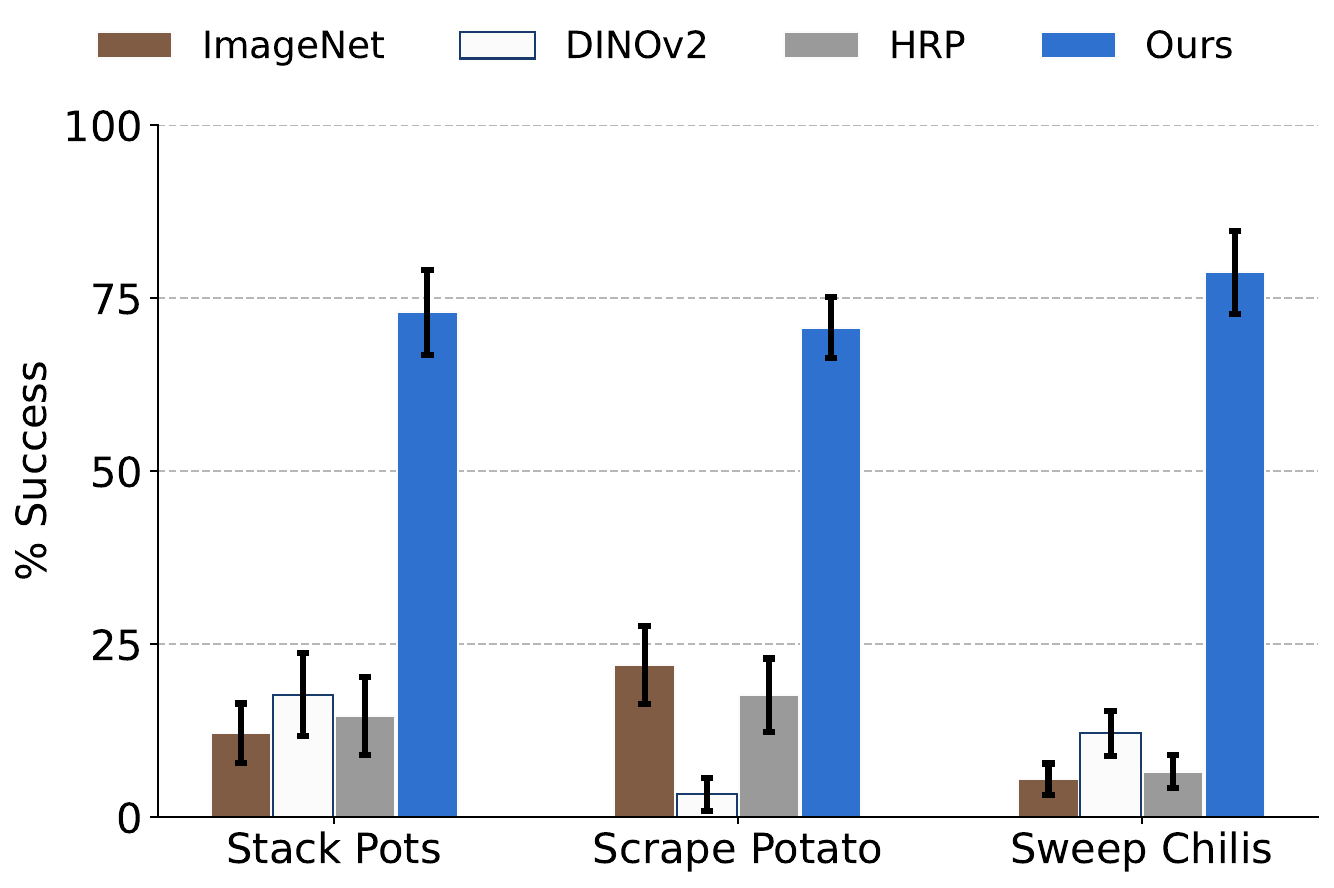}
    \caption{Average success rate (\%) on three bimanual tasks—Stack Pots, Scrape Potato, Sweep Chilis. Each task is evaluated over three out-of-distribution scenes (10 rollouts per scene, 30 per task). Our method, Masquerade, substantially outperforms all baselines; error bars show ± SEM.}
    \label{fig:results_mean}
\end{figure}

We evaluate our model on three tasks. We collect 50 robot demos in a single scene for each task, and evaluate each task in three OOD scenes. As shown in Fig. \ref{fig:results_mean}, our model strongly outperforms all baselines in every OOD scene we test by an average of 62 percentage points ($12\% \rightarrow 74\%$).

\subsection{Do robot overlays improve performance?}
\begin{figure}
    \centering
    \includegraphics[width=\linewidth]{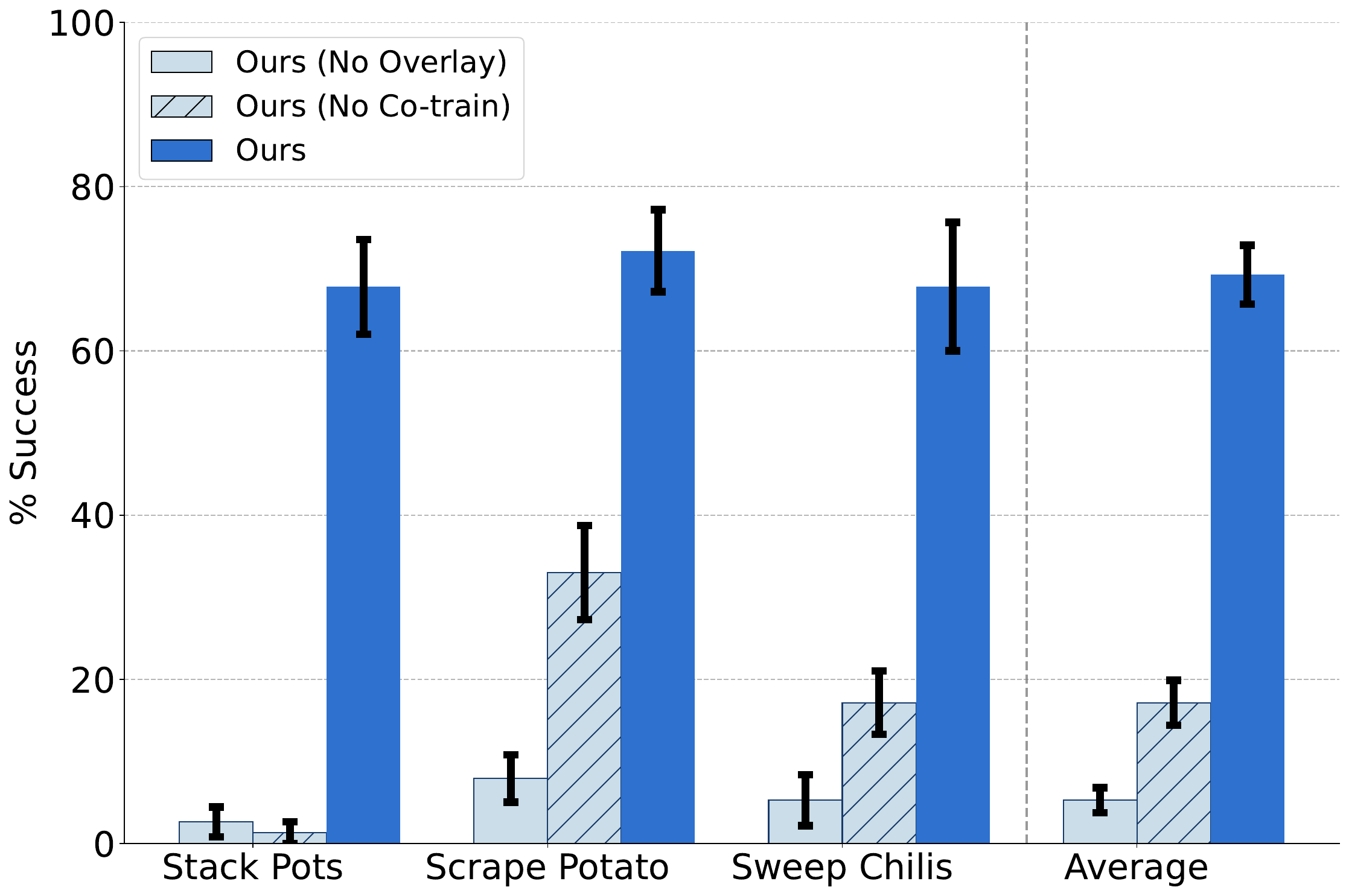}
    \caption{Ablation study on the the Stack pots, Scrape Potato and Sweep Chilis tasks demonstrating that both robot overlays and co-training are essential for achieving robust success rates in out-of-distribution settings. Results are evaluated in OOD scene 1. 25 rollouts per bar.}
    \label{fig:overlay_and_cotrain_ablation}
\end{figure}

Next, we evaluate how important editing human videos with robot overlays is to policy performance. We train a variant of our model that is pre-trained and co-trained on the same dataset our model was trained on but using raw human videos (no overlays). We evaluate our policy on all three tasks in OOD Scene 1 (Fig. \ref{fig:overlay_and_cotrain_ablation}), and find that this “no-overlay” model suffers a steep performance drop—showing that closing the embodiment gap with robot overlays unlocks far more learning from human videos.

\subsection{Does cotraining improve performance?}
We also ablate the use of co-training during policy training. We test a version of our method that first pretrains a vision encoder on edited human videos \(\mathcal{D}_{\mathrm{human}}'\) and finetunes it purely on the policy loss $\mathcal{L}
= \mathcal{L}_{\text{policy}}$. Fig. \ref{fig:overlay_and_cotrain_ablation} shows that removing co-training leads to a dramatic performance drop—demonstrating that without co-training, the encoder forgets the valuable representations learned from human videos. Co-training is therefore critical for preserving that knowledge and maintaining high task performance.

\subsection{Does increasing the amount of in-the-wild data improve performance?}
\begin{figure}
    \centering
    \includegraphics[width=\linewidth]{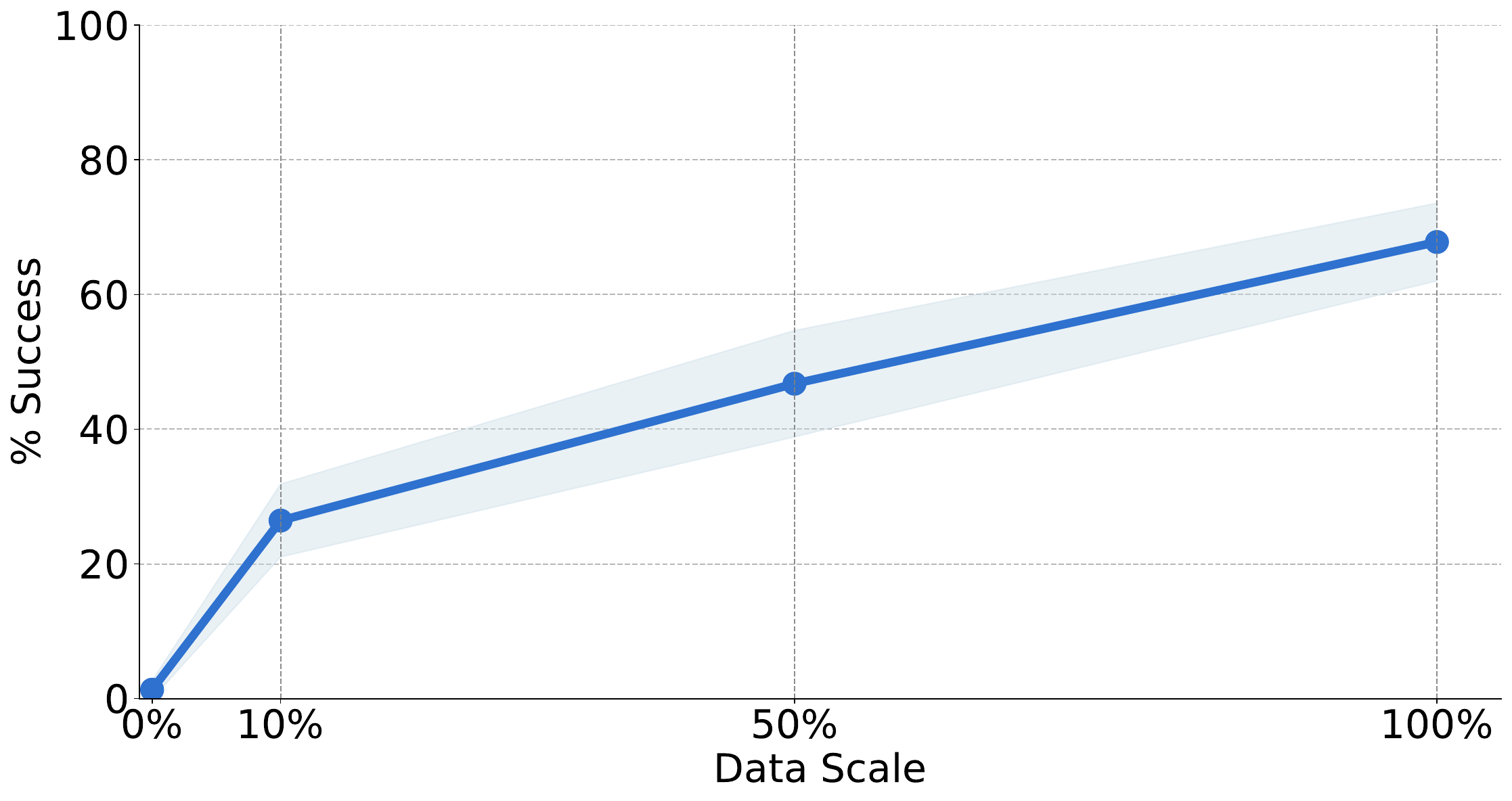}
    \caption{Data scaling experiment: Average success rate (\%) as a function of the fraction of edited human videos used during co-training (0\%, 10\%, 50\%, 100\%). Results are for the Stack Pots task in OOD scene 1. Error bars show ± SEM over 25 rollouts. Success rises monotonically with more videos—confirming that edited human-video data directly drives policy performance.}
    \label{fig:data_scale_ablation}
\end{figure}

To confirm the contribution of edited human videos to policy learning, we measured performance as a function of the amount of co-training data. We subsampled our edited video dataset at 0\% (no co-training), 10\%, 50\%, and 100\% of its full size and retrained the Stack Pots policy under identical settings, with the same number of training epochs in each case. As Fig. \ref{fig:data_scale_ablation} shows, success rates rise steadily with more human-video data: 0\% → 2\%, 10\% → 26\%, 50\% → 47\%, and 100\% → 68\% (25 rollouts each). This clear upward trend demonstrates that increasing the amount of in-the-wild human videos directly boosts robot performance and suggests further gains could be realized by scaling beyond the current dataset size.

\subsection{In-distribution vs Out-of-distribution performance}
We compare the performance of our method on the original in-distribution training scene and OOD Scene 1 for the Sweep Chilis task. Unlike all baselines, which suffer large drops, Masquerade maintains similar in-distribution and out-of-distribution performance—demonstrating its robustness to scene shifts.

\begin{figure}
    \centering
    \includegraphics[width=\linewidth]{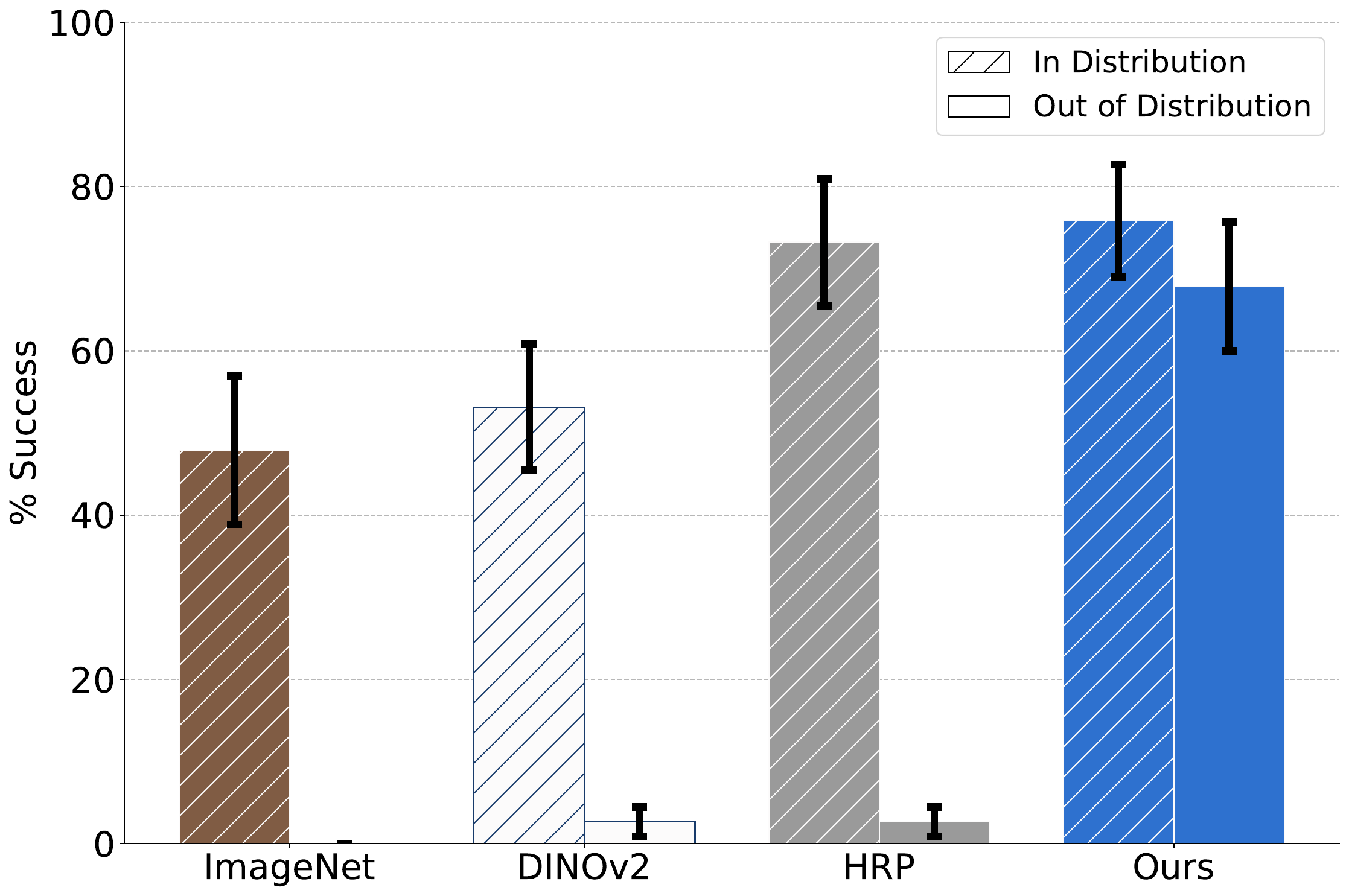}
    \caption{In‐distribution vs. out‐of‐distribution performance: average success rate (\%) for each model in the original training scene (In Distribution) and a novel scene (Out of Distribution scene 1) over 25 rollouts. Our method has the smallest drop in performance when moving to an OOD scene. Error bars show ± SEM.}
    \label{fig:id_vs_ood}
\end{figure}

\section{Limitations and Future Work}
Our approach has several limitations. First, our method relies on hand-pose estimators to align robot overlays from monocular images. These models perform poorly on frames with fast motions or heavy occlusions, and such frames must be discarded from our training dataset. However, this also means that as hand-pose estimators improve, our overlays will too. Second, the lack of depth data means that we cannot easily reason about which pixels of the robot should be overlaid on the image and which ones are actually behind objects in the scene and should therefore not be overlaid on the image. Improving this would significantly increase the realism of the grasps of our rendered robot. Third, egocentric camera motion in in-the-wild videos forces us to filter out many frames, as our method is implemented on a stationary robot without a movable camera. Improving camera pose estimation and using a mobile robot, ideally with a movable camera, could help mitigate this. Finally, because we retarget dexterous human grasps to a parallel-jaw robot, the mapping is imperfect; incorporating dexterous end-effectors and a more sophisticated retargeting pipeline would further narrow the embodiment gap.

While our work focuses on using edited human videos to improve vision representations for policy learning, the same data-editing pipeline could benefit other uses of human video in robotics, such as reward learning, motion prior extraction, or video generation. Exploring these combinations—and integrating advances in pose estimation, depth reasoning, and retargeting—offers a promising path toward scalable, web-scale robot learning from diverse, in-the-wild human videos.

\section{Conclusion}
Masquerade demonstrates that explicitly closing the visual embodiment gap between humans and robots—even via simple 2D inpainting and overlays—unlocks vast, in-the-wild human video data for policy learning. By pretraining a ViT‐Base encoder on 675K robotized frames and co‐training with only 50 real demos per task, our method achieves zero‐shot transfer to unseen scenes, outperforming baselines by over 5× on three long‐horizon bimanual tasks (Fig. \ref{fig:results_mean}) and exhibiting minimal drop from in-distribution to out-of-distribution settings (Fig. \ref{fig:id_vs_ood}). 

Ablations confirm that both the robot overlay and co‐training objectives are indispensable (Fig. \ref{fig:overlay_and_cotrain_ablation}), and scaling the human video corpus yields steadily improving success rates (Fig. \ref{fig:data_scale_ablation}), suggesting further gains with larger datasets. Future work in improving overlays, handling egocentric camera motion, and more expressive retargeting to dexterous grippers could further pave the way toward truly scalable, web-scale robot learning from human video.


%%%%%%%%%%%%%%%%%%%%%%%%%%%%%%%%%%%%%%%%%%%%%%%%%%%%%%%%%%%%%%%%%%%%%%%%%%%%%%%%

%%%%%%%%%%%%%%%%%%%%%%%%%%%%%%%%%%%%%%%%%%%%%%%%%%%%%%%%%%%%%%%%%%%%%%%%%%%%%%%%

%%%%%%%%%%%%%%%%%%%%%%%%%%%%%%%%%%%%%%%%%%%%%%%%%%%%%%%%%%%%%%%%%%%%%%%%%%%%%%%%


\section*{Acknowledgment}
This work was supported by the NSF through grant number \#2327974. We thank Jingyun Yang, Carlota Parés-Morlans, Claire Chen, Zen Yaskawa and the Stanford Robotics Center for their valuable help throughout the project.

%%%%%%%%%%%%%%%%%%%%%%%%%%%%%%%%%%%%%%%%%%%%%%%%%%%%%%%%%%%%%%%%%%%%%%%%%%%%%%%%
\bibliographystyle{IEEEtran}
\bibliography{references}  % .bib

\begin{thebibliography}{10}
\providecommand{\url}[1]{#1}
\csname url@rmstyle\endcsname
\providecommand{\newblock}{\relax}
\providecommand{\bibinfo}[2]{#2}
\providecommand\BIBentrySTDinterwordspacing{\spaceskip=0pt\relax}
\providecommand\BIBentryALTinterwordstretchfactor{4}
\providecommand\BIBentryALTinterwordspacing{\spaceskip=\fontdimen2\font plus
\BIBentryALTinterwordstretchfactor\fontdimen3\font minus \fontdimen4\font\relax}
\providecommand\BIBforeignlanguage[2]{{%
\expandafter\ifx\csname l@#1\endcsname\relax
\typeout{** WARNING: IEEEtran.bst: No hyphenation pattern has been}%
\typeout{** loaded for the language `#1'. Using the pattern for}%
\typeout{** the default language instead.}%
\else
\language=\csname l@#1\endcsname
\fi
#2}}

\bibitem{nair2023r3m}
S.~Nair, A.~Rajeswaran, V.~Kumar, C.~Finn, and A.~Gupta, ``R3m: A universal visual representation for robot manipulation,'' in \emph{Proceedings of The 6th Conference on Robot Learning}, ser. Proceedings of Machine Learning Research, K.~Liu, D.~Kulic, and J.~Ichnowski, Eds., vol. 205.\hskip 1em plus 0.5em minus 0.4em\relax PMLR, 14--18 Dec 2023, pp. 892--909.

\bibitem{karamcheti2023voltron}
S.~Karamcheti, S.~Nair, A.~S. Chen, T.~Kollar, C.~Finn, D.~Sadigh, and P.~Liang, ``Language-driven representation learning for robotics,'' in \emph{Robotics: Science and Systems XIX, Daegu, Republic of Korea, July 10-14, 2023}, K.~E. Bekris, K.~Hauser, S.~L. Herbert, and J.~Yu, Eds., 2023.

\bibitem{radosavovic2023real}
I.~Radosavovic, T.~Xiao, S.~James, P.~Abbeel, J.~Malik, and T.~Darrell, ``Real-world robot learning with masked visual pre-training,'' in \emph{Proceedings of The 6th Conference on Robot Learning}, ser. Proceedings of Machine Learning Research, K.~Liu, D.~Kulic, and J.~Ichnowski, Eds., vol. 205.\hskip 1em plus 0.5em minus 0.4em\relax PMLR, 14--18 Dec 2023, pp. 416--426.

\bibitem{majumdar2023we}
A.~Majumdar, K.~Yadav, S.~Arnaud, Y.~J. Ma, C.~Chen, S.~Silwal, A.~Jain, V.~Berges, T.~Wu, J.~Vakil, P.~Abbeel, J.~Malik, D.~Batra, Y.~Lin, O.~Maksymets, A.~Rajeswaran, and F.~Meier, ``Where are we in the search for an artificial visual cortex for embodied intelligence?'' in \emph{Advances in Neural Information Processing Systems 36: Annual Conference on Neural Information Processing Systems 2023, NeurIPS 2023, New Orleans, LA, USA, December 10 - 16, 2023}, A.~Oh, T.~Naumann, A.~Globerson, K.~Saenko, M.~Hardt, and S.~Levine, Eds., 2023.

\bibitem{srirama2024hrp}
M.~K. Srirama, S.~Dasari, S.~Bahl, and A.~Gupta, ``{HRP:} human affordances for robotic pre-training,'' in \emph{Robotics: Science and Systems XX, Delft, The Netherlands, July 15-19, 2024}, D.~Kulic, G.~Venture, K.~E. Bekris, and E.~Coronado, Eds., 2024.

\bibitem{chen2021learning}
A.~S. Chen, S.~Nair, and C.~Finn, ``{Learning Generalizable Robotic Reward Functions from “In-The-Wild” Human Videos},'' in \emph{Proceedings of Robotics: Science and Systems}, Virtual, July 2021.

\bibitem{mavip}
Y.~J. Ma, S.~Sodhani, D.~Jayaraman, O.~Bastani, V.~Kumar, and A.~Zhang, ``{VIP:} towards universal visual reward and representation via value-implicit pre-training,'' in \emph{The Eleventh International Conference on Learning Representations, {ICLR} 2023, Kigali, Rwanda, May 1-5, 2023}.\hskip 1em plus 0.5em minus 0.4em\relax OpenReview.net, 2023.

\bibitem{bhateja2024robotic}
C.~Bhateja, D.~Guo, D.~Ghosh, A.~Singh, M.~Tomar, Q.~Vuong, Y.~Chebotar, S.~Levine, and A.~Kumar, ``Robotic offline rl from internet videos via value-function learning,'' in \emph{2024 IEEE International Conference on Robotics and Automation (ICRA)}, 2024, pp. 16\,977--16\,984.

\bibitem{wu2023unleashing}
H.~Wu, Y.~Jing, C.~Cheang, G.~Chen, J.~Xu, X.~Li, M.~Liu, H.~Li, and T.~Kong, ``Unleashing large-scale video generative pre-training for visual robot manipulation,'' in \emph{The Twelfth International Conference on Learning Representations, {ICLR} 2024, Vienna, Austria, May 7-11, 2024}.\hskip 1em plus 0.5em minus 0.4em\relax OpenReview.net, 2024.

\bibitem{du2023learning}
Y.~Du, S.~Yang, B.~Dai, H.~Dai, O.~Nachum, J.~Tenenbaum, D.~Schuurmans, and P.~Abbeel, ``Learning universal policies via text-guided video generation,'' in \emph{Advances in Neural Information Processing Systems 36: Annual Conference on Neural Information Processing Systems 2023, NeurIPS 2023, New Orleans, LA, USA, December 10 - 16, 2023}, A.~Oh, T.~Naumann, A.~Globerson, K.~Saenko, M.~Hardt, and S.~Levine, Eds., 2023.

\bibitem{bharadhwaj2024gen2act}
H.~Bharadhwaj, D.~Dwibedi, A.~Gupta, S.~Tulsiani, C.~Doersch, T.~Xiao, D.~Shah, F.~Xia, D.~Sadigh, and S.~Kirmani, ``Gen2act: Human video generation in novel scenarios enables generalizable robot manipulation,'' \emph{arXiv preprint arXiv:2409.16283}, 2024.

\bibitem{cheang2024gr}
C.-L. Cheang, G.~Chen, Y.~Jing, T.~Kong, H.~Li, Y.~Li, Y.~Liu, H.~Wu, J.~Xu, Y.~Yang, \emph{et~al.}, ``Gr-2: A generative video-language-action model with web-scale knowledge for robot manipulation,'' \emph{arXiv preprint arXiv:2410.06158}, 2024.

\bibitem{lepert2025phantom}
M.~Lepert, J.~Fang, and J.~Bohg, ``Phantom: Training robots without robots using only human videos,'' \emph{arXiv preprint arXiv:2503.00779}, 2025.

\bibitem{grauman2022ego4d}
K.~Grauman, A.~Westbury, E.~Byrne, Z.~Chavis, A.~Furnari, R.~Girdhar, J.~Hamburger, H.~Jiang, M.~Liu, X.~Liu, M.~Martin, T.~Nagarajan, I.~Radosavovic, S.~K. Ramakrishnan, F.~Ryan, J.~Sharma, M.~Wray, M.~Xu, E.~Z. Xu, C.~Zhao, S.~Bansal, D.~Batra, V.~Cartillier, S.~Crane, T.~Do, M.~Doulaty, A.~Erapalli, C.~Feichtenhofer, A.~Fragomeni, Q.~Fu, A.~Gebreselasie, C.~Gonz{\'{a}}lez, J.~Hillis, X.~Huang, Y.~Huang, W.~Jia, W.~Khoo, J.~Kol{\'{a}}r, S.~Kottur, A.~Kumar, F.~Landini, C.~Li, Y.~Li, Z.~Li, K.~Mangalam, R.~Modhugu, J.~Munro, T.~Murrell, T.~Nishiyasu, W.~Price, P.~R. Puentes, M.~Ramazanova, L.~Sari, K.~K. Somasundaram, A.~Southerland, Y.~Sugano, R.~Tao, M.~Vo, Y.~Wang, X.~Wu, T.~Yagi, Z.~Zhao, Y.~Zhu, P.~Arbel{\'{a}}ez, D.~Crandall, D.~Damen, G.~M. Farinella, C.~Fuegen, B.~Ghanem, V.~K. Ithapu, C.~V. Jawahar, H.~Joo, K.~Kitani, H.~Li, R.~A. Newcombe, A.~Oliva, H.~S. Park, J.~M. Rehg, Y.~Sato, J.~Shi, M.~Z. Shou, A.~Torralba, L.~Torresani, M.~Yan, and J.~Malik, ``Ego4d: Around the world in 3, 000 hours of
  egocentric video,'' in \emph{{IEEE/CVF} Conference on Computer Vision and Pattern Recognition, {CVPR} 2022, New Orleans, LA, USA, June 18-24, 2022}.\hskip 1em plus 0.5em minus 0.4em\relax {IEEE}, 2022, pp. 18\,973--18\,990.

\bibitem{he2022masked}
K.~He, X.~Chen, S.~Xie, Y.~Li, P.~Doll{\'{a}}r, and R.~B. Girshick, ``Masked autoencoders are scalable vision learners,'' in \emph{{IEEE/CVF} Conference on Computer Vision and Pattern Recognition, {CVPR} 2022, New Orleans, LA, USA, June 18-24, 2022}.\hskip 1em plus 0.5em minus 0.4em\relax {IEEE}, 2022, pp. 15\,979--15\,988.

\bibitem{Shao-RSS-20}
L.~Shao, T.~Migimatsu, Q.~Zhang, K.~Yang, and J.~Bohg, ``{Concept2Robot: Learning Manipulation Concepts from Instructions and Human Demonstrations},'' in \emph{Proceedings of Robotics: Science and Systems}, Corvalis, Oregon, USA, July 2020.

\bibitem{shaw2023videodex}
K.~Shaw, S.~Bahl, and D.~Pathak, ``Videodex: Learning dexterity from internet videos,'' in \emph{Proceedings of The 6th Conference on Robot Learning}, ser. Proceedings of Machine Learning Research, K.~Liu, D.~Kulic, and J.~Ichnowski, Eds., vol. 205.\hskip 1em plus 0.5em minus 0.4em\relax PMLR, 14--18 Dec 2023, pp. 654--665.

\bibitem{mandikal2022dexvip}
P.~Mandikal and K.~Grauman, ``Dexvip: Learning dexterous grasping with human hand pose priors from video,'' in \emph{Proceedings of the 5th Conference on Robot Learning}, ser. Proceedings of Machine Learning Research, A.~Faust, D.~Hsu, and G.~Neumann, Eds., vol. 164.\hskip 1em plus 0.5em minus 0.4em\relax PMLR, 08--11 Nov 2022, pp. 651--661.

\bibitem{singh2024hand}
H.~G. Singh, A.~Loquercio, C.~Sferrazza, J.~Wu, H.~Qi, P.~Abbeel, and J.~Malik, ``Hand-object interaction pretraining from videos,'' \emph{arXiv preprint arXiv:2409.08273}, 2024.

\bibitem{kannan2023deft}
A.~Kannan, K.~Shaw, S.~Bahl, P.~Mannam, and D.~Pathak, ``Deft: Dexterous fine-tuning for hand policies,'' in \emph{Proceedings of The 7th Conference on Robot Learning}, ser. Proceedings of Machine Learning Research, J.~Tan, M.~Toussaint, and K.~Darvish, Eds., vol. 229.\hskip 1em plus 0.5em minus 0.4em\relax PMLR, 06--09 Nov 2023, pp. 928--942.

\bibitem{bharadhwaj2024towards}
H.~Bharadhwaj, A.~Gupta, V.~Kumar, and S.~Tulsiani, ``Towards generalizable zero-shot manipulation via translating human interaction plans,'' in \emph{2024 IEEE International Conference on Robotics and Automation (ICRA)}, 2024, pp. 6904--6911.

\bibitem{bharadhwaj2024track2act}
H.~Bharadhwaj, R.~Mottaghi, A.~Gupta, and S.~Tulsiani, ``Track2act: Predicting point tracks from internet videos enables generalizable robot manipulation,'' in \emph{Computer Vision - {ECCV} 2024 - 18th European Conference, Milan, Italy, September 29-October 4, 2024, Proceedings, Part {LXXVI}}, ser. Lecture Notes in Computer Science, A.~Leonardis, E.~Ricci, S.~Roth, O.~Russakovsky, T.~Sattler, and G.~Varol, Eds., vol. 15134.\hskip 1em plus 0.5em minus 0.4em\relax Springer, 2024, pp. 306--324.

\bibitem{bahl2023affordances}
S.~Bahl, R.~Mendonca, L.~Chen, U.~Jain, and D.~Pathak, ``Affordances from human videos as a versatile representation for robotics,'' in \emph{{IEEE/CVF} Conference on Computer Vision and Pattern Recognition, {CVPR} 2023, Vancouver, BC, Canada, June 17-24, 2023}.\hskip 1em plus 0.5em minus 0.4em\relax {IEEE}, 2023, pp. 1--13.

\bibitem{chen2025vidbot}
H.~Chen, B.~Sun, A.~Zhang, M.~Pollefeys, and S.~Leutenegger, ``Vidbot: Learning generalizable 3d actions from in-the-wild 2d human videos for zero-shot robotic manipulation,'' in \emph{{IEEE/CVF} Conference on Computer Vision and Pattern Recognition, {CVPR} 2025, Nashville, TN, USA, June 11-15, 2025}.\hskip 1em plus 0.5em minus 0.4em\relax Computer Vision Foundation / {IEEE}, 2025, pp. 27\,661--27\,672.

\bibitem{shi2025zeromimic}
J.~Shi, Z.~Zhao, T.~Wang, I.~Pedroza, A.~Luo, J.~Wang, J.~Ma, and D.~Jayaraman, ``Zeromimic: Distilling robotic manipulation skills from web videos,'' in \emph{International Conference on Robotics and Automation (ICRA)}, 2025.

\bibitem{yelatent}
S.~Ye, J.~Jang, B.~Jeon, S.~J. Joo, J.~Yang, B.~Peng, A.~Mandlekar, R.~Tan, Y.~Chao, B.~Y. Lin, L.~Liden, K.~Lee, J.~Gao, L.~Zettlemoyer, D.~Fox, and M.~Seo, ``Latent action pretraining from videos,'' in \emph{The Thirteenth International Conference on Learning Representations, {ICLR} 2025, Singapore, April 24-28, 2025}.\hskip 1em plus 0.5em minus 0.4em\relax OpenReview.net, 2025.

\bibitem{van2017neural}
A.~van~den Oord, O.~Vinyals, and K.~Kavukcuoglu, ``Neural discrete representation learning,'' in \emph{Advances in Neural Information Processing Systems 30: Annual Conference on Neural Information Processing Systems 2017, December 4-9, 2017, Long Beach, CA, {USA}}, I.~Guyon, U.~von Luxburg, S.~Bengio, H.~M. Wallach, R.~Fergus, S.~V.~N. Vishwanathan, and R.~Garnett, Eds., 2017, pp. 6306--6315.

\bibitem{li2025h2r}
G.~Li, Y.~Lyu, Z.~Liu, C.~Hou, J.~Zhang, and S.~Zhang, ``H2r: A human-to-robot data augmentation for robot pre-training from videos,'' \emph{arXiv preprint arXiv:2505.11920}, 2025.

\bibitem{kareer2024egomimic}
S.~Kareer, D.~Patel, R.~Punamiya, P.~Mathur, S.~Cheng, C.~Wang, J.~Hoffman, and D.~Xu, ``Egomimic: Scaling imitation learning via egocentric video,'' in \emph{1st Workshop on X-Embodiment Robot Learning}, 2024.

\bibitem{qiu2025humanoid}
R.-Z. Qiu, S.~Yang, X.~Cheng, C.~Chawla, J.~Li, T.~He, G.~Yan, D.~J. Yoon, R.~Hoque, L.~Paulsen, \emph{et~al.}, ``Humanoid policy\~{} human policy,'' \emph{arXiv preprint arXiv:2503.13441}, 2025.

\bibitem{yang2025egovla}
R.~Yang, Q.~Yu, Y.~Wu, R.~Yan, B.~Li, A.-C. Cheng, X.~Zou, Y.~Fang, H.~Yin, S.~Liu, \emph{et~al.}, ``Egovla: Learning vision-language-action models from egocentric human videos,'' \emph{arXiv preprint arXiv:2507.12440}, 2025.

\bibitem{zakka2022xirl}
K.~Zakka, A.~Zeng, P.~Florence, J.~Tompson, J.~Bohg, and D.~Dwibedi, ``Xirl: Cross-embodiment inverse reinforcement learning,'' in \emph{Proceedings of the 5th Conference on Robot Learning}, ser. Proceedings of Machine Learning Research, A.~Faust, D.~Hsu, and G.~Neumann, Eds., vol. 164.\hskip 1em plus 0.5em minus 0.4em\relax PMLR, 08--11 Nov 2022, pp. 537--546.

\bibitem{wang2023mimicplay}
C.~Wang, L.~Fan, J.~Sun, R.~Zhang, L.~Fei-Fei, D.~Xu, Y.~Zhu, and A.~Anandkumar, ``Mimicplay: Long-horizon imitation learning by watching human play,'' in \emph{Proceedings of The 7th Conference on Robot Learning}, ser. Proceedings of Machine Learning Research, J.~Tan, M.~Toussaint, and K.~Darvish, Eds., vol. 229.\hskip 1em plus 0.5em minus 0.4em\relax PMLR, 06--09 Nov 2023, pp. 201--221.

\bibitem{xu2023xskill}
M.~Xu, Z.~Xu, C.~Chi, M.~Veloso, and S.~Song, ``Xskill: Cross embodiment skill discovery,'' in \emph{Proceedings of The 7th Conference on Robot Learning}, ser. Proceedings of Machine Learning Research, J.~Tan, M.~Toussaint, and K.~Darvish, Eds., vol. 229.\hskip 1em plus 0.5em minus 0.4em\relax PMLR, 06--09 Nov 2023, pp. 3536--3555.

\bibitem{qin2022dexmv}
Y.~Qin, Y.-H. Wu, S.~Liu, H.~Jiang, R.~Yang, Y.~Fu, and X.~Wang, ``Dexmv: Imitation learning for dexterous manipulation from human videos,'' in \emph{European Conference on Computer Vision}.\hskip 1em plus 0.5em minus 0.4em\relax Springer, 2022, pp. 570--587.

\bibitem{zhu2024vision}
Y.~Zhu, A.~Lim, P.~Stone, and Y.~Zhu, ``Vision-based manipulation from single human video with open-world object graphs,'' in \emph{1st Workshop on X-Embodiment Robot Learning}, 2024.

\bibitem{heppert2024ditto}
N.~Heppert, M.~Argus, T.~Welschehold, T.~Brox, and A.~Valada, ``Ditto: Demonstration imitation by trajectory transformation,'' in \emph{2024 IEEE/RSJ International Conference on Intelligent Robots and Systems (IROS)}, 2024, pp. 7565--7572.

\bibitem{xu2024flow}
M.~Xu, Z.~Xu, Y.~Xu, C.~Chi, G.~Wetzstein, M.~Veloso, and S.~Song, ``Flow as the cross-domain manipulation interface,'' in \emph{Proceedings of The 8th Conference on Robot Learning}, ser. Proceedings of Machine Learning Research, P.~Agrawal, O.~Kroemer, and W.~Burgard, Eds., vol. 270.\hskip 1em plus 0.5em minus 0.4em\relax PMLR, 06--09 Nov 2024, pp. 2475--2499.

\bibitem{haldar2025point}
S.~Haldar and L.~Pinto, ``Point policy: Unifying observations and actions with key points for robot manipulation,'' \emph{arXiv preprint arXiv:2502.20391}, 2025.

\bibitem{ren2025motion}
J.~Ren, P.~Sundaresan, D.~Sadigh, S.~Choudhury, and J.~Bohg, ``Motion tracks: A unified representation for human-robot transfer in few-shot imitation learning,'' 2025.

\bibitem{lum2025crossinghumanrobotembodimentgap}
\BIBentryALTinterwordspacing
T.~G.~W. Lum, O.~Y. Lee, C.~K. Liu, and J.~Bohg, ``Crossing the human-robot embodiment gap with sim-to-real rl using one human demonstration,'' 2025. [Online]. Available: \url{https://arxiv.org/abs/2504.12609}
\BIBentrySTDinterwordspacing

\bibitem{bahl2022human}
S.~Bahl, A.~Gupta, and D.~Pathak, ``Human-to-robot imitation in the wild,'' in \emph{Robotics: Science and Systems XVIII, New York City, NY, USA, June 27 - July 1, 2022}, K.~Hauser, D.~A. Shell, and S.~Huang, Eds., 2022.

\bibitem{duan2023ar2}
J.~Duan, Y.~R. Wang, M.~Shridhar, D.~Fox, and R.~Krishna, ``Ar2-d2: Training a robot without a robot,'' in \emph{Proceedings of The 7th Conference on Robot Learning}, ser. Proceedings of Machine Learning Research, J.~Tan, M.~Toussaint, and K.~Darvish, Eds., vol. 229.\hskip 1em plus 0.5em minus 0.4em\relax PMLR, 06--09 Nov 2023, pp. 2838--2848.

\bibitem{Damen2018EPICKITCHENS}
D.~Damen, H.~Doughty, G.~M. Farinella, S.~Fidler, A.~Furnari, E.~Kazakos, D.~Moltisanti, J.~Munro, T.~Perrett, W.~Price, \emph{et~al.}, ``Scaling egocentric vision: The epic-kitchens dataset,'' in \emph{Proceedings of the European conference on computer vision (ECCV)}, 2018, pp. 720--736.

\bibitem{pavlakos2024reconstructing}
G.~Pavlakos, D.~Shan, I.~Radosavovic, A.~Kanazawa, D.~Fouhey, and J.~Malik, ``Reconstructing hands in 3d with transformers,'' in \emph{{IEEE/CVF} Conference on Computer Vision and Pattern Recognition, {CVPR} 2024, Seattle, WA, USA, June 16-22, 2024}.\hskip 1em plus 0.5em minus 0.4em\relax {IEEE}, 2024, pp. 9826--9836.

\bibitem{wu2019detectron2}
Y.~Wu, A.~Kirillov, F.~Massa, W.-Y. Lo, and R.~Girshick, ``Detectron2,'' \url{https://github.com/facebookresearch/detectron2}, 2019.

\bibitem{ravisam}
N.~Ravi, V.~Gabeur, Y.~Hu, R.~Hu, C.~Ryali, T.~Ma, H.~Khedr, R.~R{\"{a}}dle, C.~Rolland, L.~Gustafson, E.~Mintun, J.~Pan, K.~V. Alwala, N.~Carion, C.~Wu, R.~B. Girshick, P.~Doll{\'{a}}r, and C.~Feichtenhofer, ``{SAM} 2: Segment anything in images and videos,'' in \emph{The Thirteenth International Conference on Learning Representations, {ICLR} 2025, Singapore, April 24-28, 2025}.\hskip 1em plus 0.5em minus 0.4em\relax OpenReview.net, 2025.

\bibitem{liCvpr22vInpainting}
Z.~Li, C.~Lu, J.~Qin, C.~Guo, and M.~Cheng, ``Towards an end-to-end framework for flow-guided video inpainting,'' in \emph{{IEEE/CVF} Conference on Computer Vision and Pattern Recognition, {CVPR} 2022, New Orleans, LA, USA, June 18-24, 2022}.\hskip 1em plus 0.5em minus 0.4em\relax {IEEE}, 2022, pp. 17\,541--17\,550.

\bibitem{perez2018film}
E.~Perez, F.~Strub, H.~de~Vries, V.~Dumoulin, and A.~C. Courville, ``Film: Visual reasoning with a general conditioning layer,'' in \emph{Proceedings of the Thirty-Second {AAAI} Conference on Artificial Intelligence, (AAAI-18), the 30th innovative Applications of Artificial Intelligence (IAAI-18), and the 8th {AAAI} Symposium on Educational Advances in Artificial Intelligence (EAAI-18), New Orleans, Louisiana, USA, February 2-7, 2018}, S.~A. McIlraith and K.~Q. Weinberger, Eds.\hskip 1em plus 0.5em minus 0.4em\relax {AAAI} Press, 2018, pp. 3942--3951.

\bibitem{sanh2019distilbert}
V.~Sanh, L.~Debut, J.~Chaumond, and T.~Wolf, ``Distilbert, a distilled version of bert: smaller, faster, cheaper and lighter,'' \emph{arXiv preprint arXiv:1910.01108}, 2019.

\bibitem{dosovitskiy2020image}
A.~Dosovitskiy, L.~Beyer, A.~Kolesnikov, D.~Weissenborn, X.~Zhai, T.~Unterthiner, M.~Dehghani, M.~Minderer, G.~Heigold, S.~Gelly, J.~Uszkoreit, and N.~Houlsby, ``An image is worth 16x16 words: Transformers for image recognition at scale,'' in \emph{9th International Conference on Learning Representations, {ICLR} 2021, Virtual Event, Austria, May 3-7, 2021}.\hskip 1em plus 0.5em minus 0.4em\relax OpenReview.net, 2021.

\bibitem{imagenet15russakovsky}
O.~Russakovsky, J.~Deng, H.~Su, J.~Krause, S.~Satheesh, S.~Ma, Z.~Huang, A.~Karpathy, A.~Khosla, M.~S. Bernstein, A.~C. Berg, and L.~Fei{-}Fei, ``Imagenet large scale visual recognition challenge,'' \emph{Int. J. Comput. Vis.}, vol. 115, no.~3, pp. 211--252, 2015.

\bibitem{chi2023diffusionpolicy}
C.~Chi, S.~Feng, Y.~Du, Z.~Xu, E.~Cousineau, B.~Burchfiel, and S.~Song, ``Diffusion policy: Visuomotor policy learning via action diffusion,'' in \emph{Robotics: Science and Systems XIX, Daegu, Republic of Korea, July 10-14, 2023}, K.~E. Bekris, K.~Hauser, S.~L. Herbert, and J.~Yu, Eds., 2023.

\bibitem{dasari2023datasets}
S.~Dasari, M.~K. Srirama, U.~Jain, and A.~Gupta, ``An unbiased look at datasets for visuo-motor pre-training,'' in \emph{Proceedings of The 7th Conference on Robot Learning}, ser. Proceedings of Machine Learning Research, J.~Tan, M.~Toussaint, and K.~Darvish, Eds., vol. 229.\hskip 1em plus 0.5em minus 0.4em\relax PMLR, 06--09 Nov 2023, pp. 1183--1198.

\bibitem{oquab2023dinov2}
M.~Oquab, T.~Darcet, T.~Moutakanni, H.~Vo, M.~Szafraniec, V.~Khalidov, P.~Fernandez, D.~Haziza, F.~Massa, A.~El-Nouby, \emph{et~al.}, ``Dinov2: Learning robust visual features without supervision,'' \emph{arXiv preprint arXiv:2304.07193}, 2023.

\bibitem{kim2024openvla}
M.~J. Kim, K.~Pertsch, S.~Karamcheti, T.~Xiao, A.~Balakrishna, S.~Nair, R.~Rafailov, E.~P. Foster, P.~R. Sanketi, Q.~Vuong, T.~Kollar, B.~Burchfiel, R.~Tedrake, D.~Sadigh, S.~Levine, P.~Liang, and C.~Finn, ``Openvla: An open-source vision-language-action model,'' in \emph{Proceedings of The 8th Conference on Robot Learning}, ser. Proceedings of Machine Learning Research, P.~Agrawal, O.~Kroemer, and W.~Burgard, Eds., vol. 270.\hskip 1em plus 0.5em minus 0.4em\relax PMLR, 06--09 Nov 2024, pp. 2679--2713.

\bibitem{bu2025univla}
Q.~Bu, Y.~Yang, J.~Cai, S.~Gao, G.~Ren, M.~Yao, P.~Luo, and H.~Li, ``Univla: Learning to act anywhere with task-centric latent actions,'' \emph{arXiv preprint arXiv:2505.06111}, 2025.

\bibitem{lin2024data}
F.~Lin, Y.~Hu, P.~Sheng, C.~Wen, J.~You, and Y.~Gao, ``Data scaling laws in imitation learning for robotic manipulation,'' in \emph{The Thirteenth International Conference on Learning Representations, {ICLR} 2025, Singapore, April 24-28, 2025}.\hskip 1em plus 0.5em minus 0.4em\relax OpenReview.net, 2025.

\bibitem{kerr2024robot}
J.~Kerr, C.~M. Kim, M.~Wu, B.~Yi, Q.~Wang, K.~Goldberg, and A.~Kanazawa, ``Robot see robot do: Imitating articulated object manipulation with monocular 4d reconstruction,'' in \emph{Proceedings of The 8th Conference on Robot Learning}, ser. Proceedings of Machine Learning Research, P.~Agrawal, O.~Kroemer, and W.~Burgard, Eds., vol. 270.\hskip 1em plus 0.5em minus 0.4em\relax PMLR, 06--09 Nov 2024, pp. 587--603.

\bibitem{xia2024cage}
S.~Xia, H.~Fang, C.~Lu, and H.-S. Fang, ``Cage: Causal attention enables data-efficient generalizable robotic manipulation,'' \emph{arXiv preprint arXiv:2410.14974}, 2024.

\bibitem{yang2024harmonic}
R.~Yang, Y.~Kim, R.~Hendrix, A.~Kembhavi, X.~Wang, and K.~Ehsani, ``Harmonic mobile manipulation,'' in \emph{2024 IEEE/RSJ International Conference on Intelligent Robots and Systems (IROS)}, 2024, pp. 3658--3665.

\bibitem{fang2025airexo}
H.~Fang, C.~Wang, Y.~Wang, J.~Chen, S.~Xia, J.~Lv, Z.~He, X.~Yi, Y.~Guo, X.~Zhan, \emph{et~al.}, ``Airexo-2: Scaling up generalizable robotic imitation learning with low-cost exoskeletons,'' in \emph{7th Robot Learning Workshop: Towards Robots with Human-Level Abilities}, 2025.

\bibitem{huang2024rekep}
W.~Huang, C.~Wang, Y.~Li, R.~Zhang, and L.~Fei-Fei, ``Rekep: Spatio-temporal reasoning of relational keypoint constraints for robotic manipulation,'' in \emph{Proceedings of The 8th Conference on Robot Learning}, ser. Proceedings of Machine Learning Research, P.~Agrawal, O.~Kroemer, and W.~Burgard, Eds., vol. 270.\hskip 1em plus 0.5em minus 0.4em\relax PMLR, 06--09 Nov 2024, pp. 4573--4602.

\bibitem{jia2024lift3d}
Y.~Jia, J.~Liu, S.~Chen, C.~Gu, Z.~Wang, L.~Luo, L.~Lee, P.~Wang, Z.~Wang, R.~Zhang, \emph{et~al.}, ``Lift3d foundation policy: Lifting 2d large-scale pretrained models for robust 3d robotic manipulation,'' \emph{arXiv preprint arXiv:2411.18623}, 2024.

\end{thebibliography}

\clearpage

\section*{Appendix}
\begin{figure*}
    \centering
    \includegraphics[width=\textwidth]{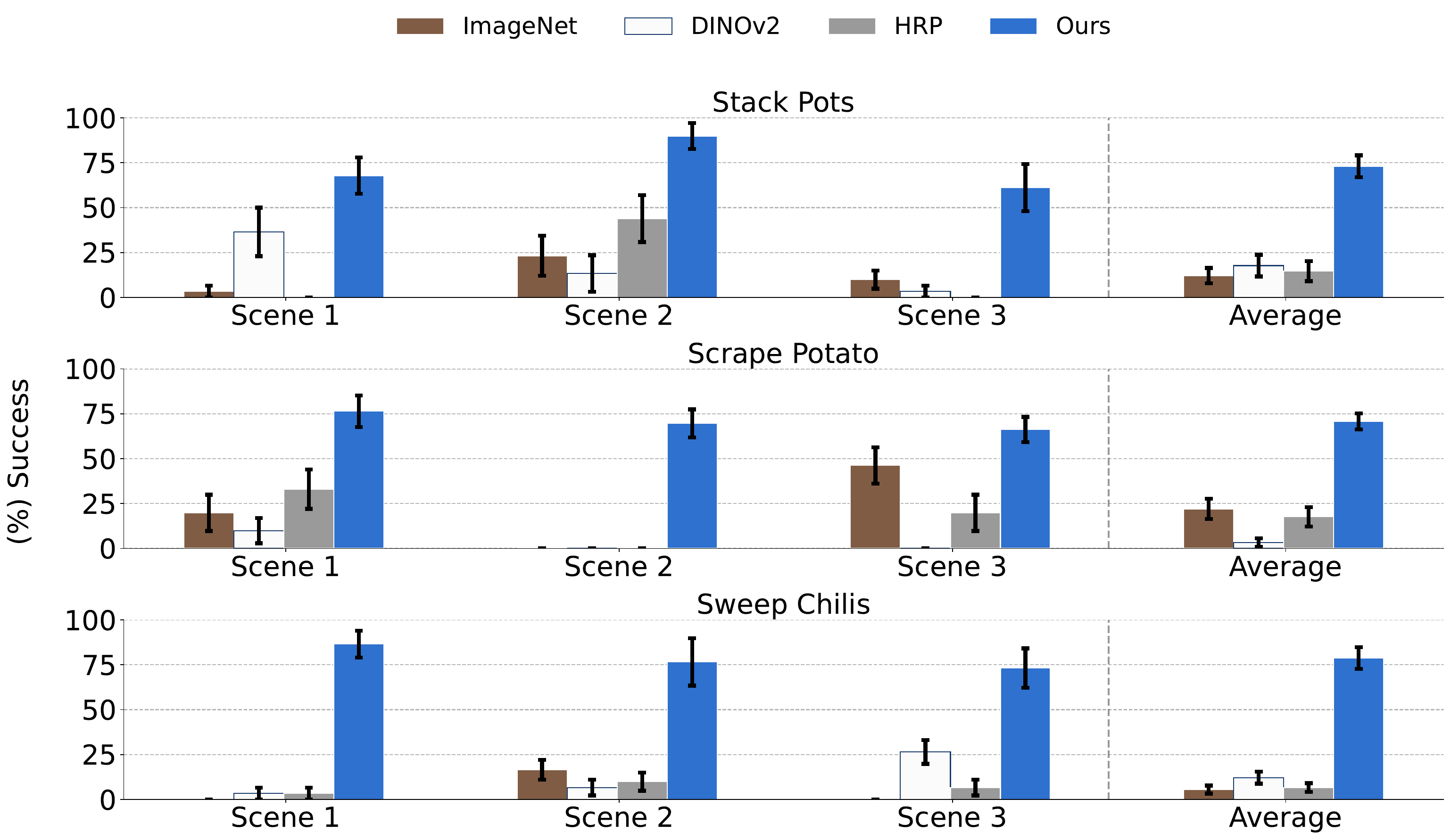}
    \caption{Average success rate (\%) on three bimanual tasks—Stack Pots, Scrape Potato, Sweep Chilis. Each task is evaluated over three out-of-distribution scenes (10 rollouts per scene, 30 per task). Our method, Masquerade, substantially outperforms all baselines; error bars show ± SEM.}
    \label{fig:results}
\end{figure*}

\subsection{Policy Training Details}

\begin{table}[h]
  \caption{Training configurations}
  \label{tab:train-details}
  \centering
  \small
  \begin{tabular}{lcc}
    \toprule
     & \textbf{Vision Encoder} & \textbf{Policy} \\
    \midrule
    \textbf{Architecture}       & ViT-Base (86 M) & Diffusion Policy \cite{chi2023diffusionpolicy} \\
    \textbf{Input Size}     & 224×224        & 224×224 \\
    \textbf{Batch size}     & 160             & 64 \\
    \textbf{LR}             & $1\times10^{-4}$      & $1\times10^{-4}$ \\
    \textbf{Optimizer}      & AdamW & AdamW \\
    \textbf{Scheduler}      & –              & Cosine (500 warmup) \\
    \textbf{Steps}          & 150 000        & 40 000 \\
    \bottomrule
  \end{tabular}
\end{table}

\medskip

The diffusion policy used the DDPM noise scheduler with 100 train and inference steps. Models were trained on NVIDIA RTX 4090 and NVIDIA A5000 GPUs.

\begin{table}[h]
  \caption{Vision encoder variants}
  \label{tab:enc-variants}
  \centering
  \small
  \begin{tabular}{lccc}
    \toprule
     & \textbf{Patch} & \textbf{Pretrain} & \textbf{Weights}\\
    \midrule
    ImageNet  & 16 & MAE  & Public \\
    DINOv2    & 14 & Feature Distillation   & Public \\
    HRP       & 16 & Aux losses   & Public \\
    Masquerade& 16 & Aux 2D loss & Ours \\
    \bottomrule
  \end{tabular}
\end{table}

For co-training, we empirically tested different $\lambda$ values ($\lambda =0.5$, $\lambda =1$, $\lambda =2$, $\lambda =10$, $\lambda =40$) and found $\lambda = 10$ to work the best.

\subsection{Dataset description}
\textbf{Human Videos}: We use edited videos from the Epic Kitchens \cite{Damen2018EPICKITCHENS} dataset to train our vision encoder. In total, we use 675,713 frames for training. 

\textbf{Robot demos}: For each task, we collect 50 bimanual robot demos using an Oculus headset.

\subsection{Additional data-editing details}

From the Epic Kitchens dataset \cite{Damen2018EPICKITCHENS}, we remove all frames where the estimated camera motion exceeds 5 cm in translation or 0.5 rad in rotation per timestep. To preserve all possible actions and maintain temporal consistency, if a single hand becomes occluded or leaves the frame, its action from the last visible frame in the episode is reused for all subsequent frames. If a hand is invisible for the entire episode, it is assigned a fixed “out-of-frame” action label. Frames in which both hands are missing are discarded.

\subsection{Hardware and controller details} 
\label{hadware}
\begin{figure}[H]
    \centering
    \includegraphics[width=0.8\linewidth]{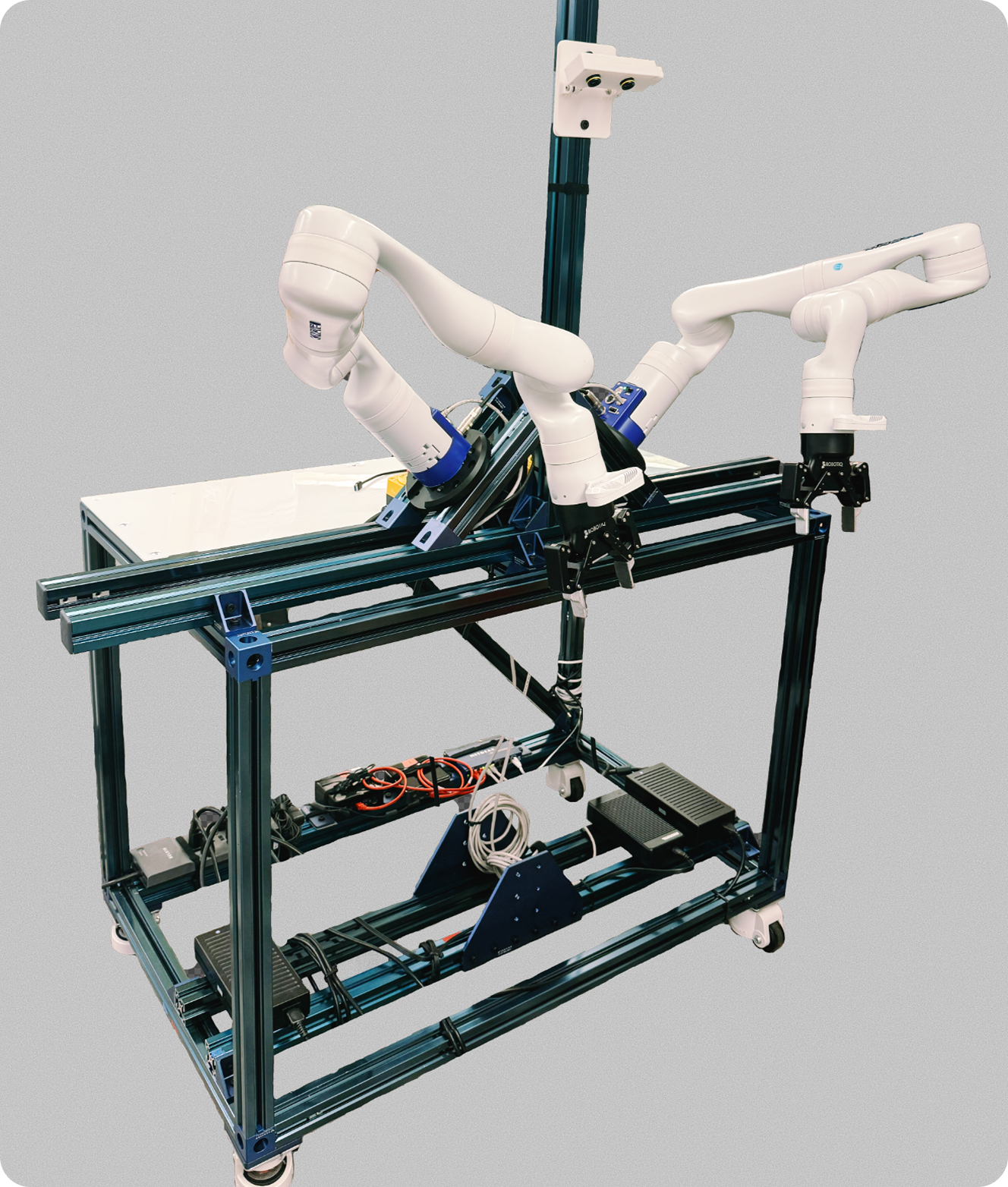}
    \caption{Our bimanual setup.}
    \label{fig:hardware}
\end{figure}

Our bimanual setup (shown in Fig. \ref{fig:hardware}) consists of two Kinova Gen3 7-dof robot arms. We control them in Cartesian space using an Inverse-Kinematics controller and a low-level joint position controller running at 1000 Hz. Each arm uses a Robotiq 2F-85 gripper (a parallel-jaw gripper) as its end-effector. A ZED mini camera with an egocentric viewpoint is rigidly mounted to our setup, providing RGB observations at each timestep. 

\subsection{Camera extrinsics detailed explanation}
Our robot's camera extrinsics are always fixed (the ZED camera is rigidly mounted), therefore the robot-camera transform is the same in every frame. For each sequence of $H$ frames in our human videos dataset, we freeze the camera pose to the first frame $t$ and interpret later keypoints ($t+1 \rightarrow t+H$) in that view via homographies. The two images below show  \textbf{start} frames from two different sequences, each rendered with the same fixed robot–camera transform. 

\begin{figure}[H]
    \centering
    \includegraphics[width=0.8\linewidth]{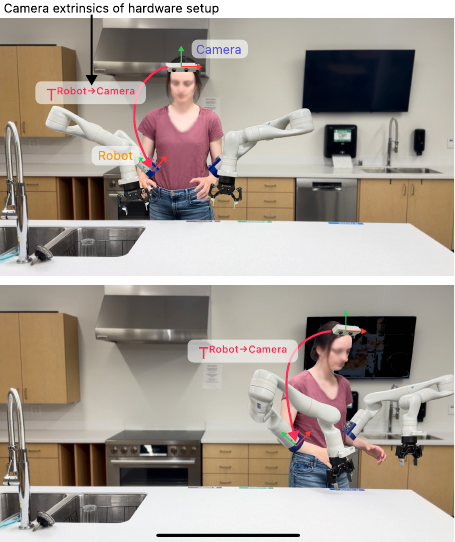}
    \caption{Two different example \textbf{start} frames showing how our robot overlays preserve a fixed robot–camera transformation within each sequence, even though the head-mounted camera may move between sequences.}
    \label{fig:camera_explanation}
\end{figure}

\subsection{Detailed scene results}
Detailed evaluation results for each task in each OOD scene are shown in Fig. \ref{fig:results}.

\subsection{Task variation description}

\begin{figure}[H]
    \centering
    \includegraphics[width=0.6\linewidth]{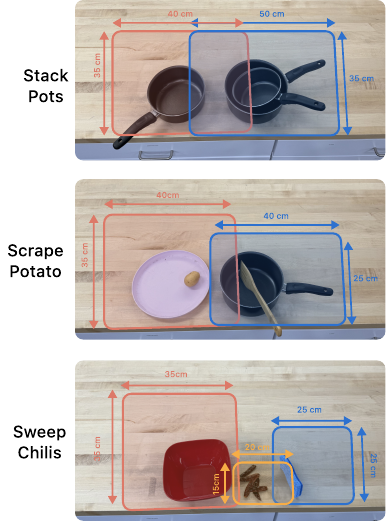}
    \caption{Variation in object placement during evaluations of each task.}
    \label{fig:task_variations}
\end{figure}

Fig. \ref{fig:task_variations}  illustrates the randomized object initialization regions (colored boxes) for each task. In Stack Pots, the left pot is placed within the yellow region, and the two right pots within the blue region. In Scrape Potato, the plate is initialized in yellow, while the pot and spatula are placed in blue. In Sweep Chilis, the red bowl starts in yellow, the sponge in blue, and the chilis in red.

\end{document}